\documentclass[letterpaper]{article}
\usepackage{aaai}
\usepackage{times}
\usepackage{helvet}
\usepackage{courier}
\usepackage{xcolor}
\usepackage{amsmath}
\usepackage{multicol,multirow}

\usepackage{algorithm,bbm,amsmath,amssymb}
\usepackage{bm}
\usepackage[noend]{algpseudocode}

\DeclareMathOperator{\kldiv}{D_{\textsc{KL}}}

\providecommand{\transprobf}[4][]{\ensuremath{\mathcal{P}({#4}{\ifthenelse{\equal{#1}{}}{}{, #1}}\mid{#2},{#3})}}

\DeclareMathOperator{\statespace}{\mathcal{S}}

\DeclareMathOperator{\actionspace}{\mathcal{A}}

\DeclareBoldMathCommand{\observation}{O}

\providecommand{\tuple}[1]{\ensuremath{\langle{#1}\rangle}}

\DeclareMathOperator*{\argmin}{arg\,min}

\usepackage{tikz}
   \usetikzlibrary{positioning,arrows,calc,trees}
   \tikzset{
   mdp/.style={>=stealth,shorten >=1pt,shorten <=1pt,auto,node distance=1.5cm,
   semithick,font=\tiny},
   state/.style={circle, draw, minimum size=0.8cm,fill=gray!15},
   transition/.style={circle, draw, minimum size=0.05cm,fill=black},
   tstate/.style={rectangle,draw,minimum size=0.8cm,fill=gray!15},
   mstate/.style={circle, draw, minimum size=0.2cm,fill=white},
   reflexive above/.style={->,loop,looseness=7,in=120,out=60},
   reflexive below/.style={->,loop,looseness=7,in=240,out=300},
   reflexive left/.style={->,loop,looseness=7,in=150,out=210},
   reflexive right/.style={->,loop,looseness=7,in=30,out=330}
   }

\makeatletter
\def\bbordermatrix#1{\begingroup \m@th
  \@tempdima 4.75\p@
  \setbox\z@\vbox{%
    \def\cr{\crcr\noalign{\kern2\p@\global\let\cr\endline}}%
    \ialign{$##$\hfil\kern2\p@\kern\@tempdima&\thinspace\hfil$##$\hfil
      &&\quad\hfil$##$\hfil\crcr
      \omit\strut\hfil\crcr\noalign{\kern-\baselineskip}%
      #1\crcr\omit\strut\cr}}%
  \setbox\tw@\vbox{\unvcopy\z@\global\setbox\@ne\lastbox}%
  \setbox\tw@\hbox{\unhbox\@ne\unskip\global\setbox\@ne\lastbox}%
  \setbox\tw@\hbox{$\kern\wd\@ne\kern-\@tempdima\left[\kern-\wd\@ne
    \global\setbox\@ne\vbox{\box\@ne\kern2\p@}%
    \vcenter{\kern-\ht\@ne\unvbox\z@\kern-\baselineskip}\,\right]$}%
  \null\;\vbox{\kern\ht\@ne\box\tw@}\endgroup}
\makeatother

\providecommand\tuple[1]{\ensuremath\langle#1\rangle}

\providecommand\initialstate{\ensuremath{s_{0}}}

\providecommand\theory{\mathbb{T}}
\providecommand\goals{\mathcal{G}}
\providecommand\goal{g}

\usepackage{graphicx}
\usepackage{hyperref}
\usepackage{nameref}

\newtheorem{definition}{Definition}

\usepackage[textsize=tiny,
	]{todonotes}

\usepackage[normalem]{ulem}

\begin{document}
%
\title{Goal Recognition as Reinforcement Learning (Preprint version)}
\author{Leonardo Amado\textsuperscript{1},
 Reuth Mirsky, \textsuperscript{2,3}, Felipe Meneguzzi \textsuperscript{4,1}\\
\textsuperscript{1} Pontifícia Universidade Católica do Rio Grande do Sul, Brazil \\
\textsuperscript{2} Bar Ilan University, Israel \\
\textsuperscript{3} The University of Texas at Austin, USA \\
\textsuperscript{4} University of Aberdeen, Scotland}

\maketitle
\begin{abstract}
\begin{quote}

Most approaches for goal recognition rely on specifications of the possible dynamics of the actor in the environment when pursuing a goal. These specifications suffer from two key issues. 
First, encoding these dynamics requires careful design by a domain expert, which is often not robust to noise at recognition time. 
Second, existing approaches often need costly real-time computations to reason about the likelihood of each potential goal.
In this paper, we develop a framework that combines model-free reinforcement learning and goal recognition to alleviate the need for careful, manual domain design, and the need for costly online executions.
This framework consists of two main stages: Offline learning of policies or utility functions for each potential goal, and online inference. 
We provide a first instance of this framework using tabular Q-learning for the learning stage, as well as three measures that can be used to perform the inference stage. 
The resulting instantiation achieves state-of-the-art performance against goal recognizers on standard evaluation domains and superior performance in noisy environments.
\end{quote}
\end{abstract}

\section{Introduction}
\label{sec:intro}
\begin{figure*}
    \centering
    \includegraphics[width=\textwidth]{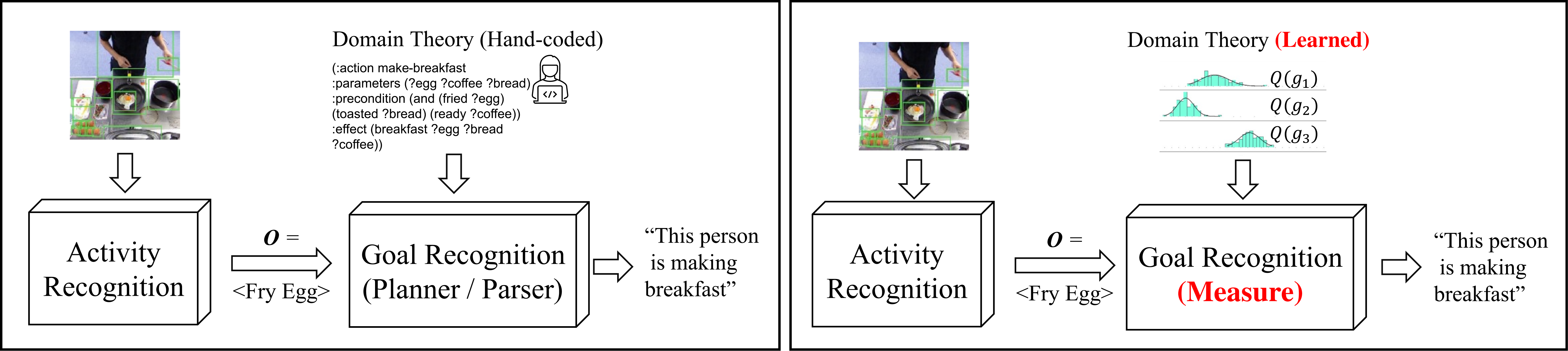}
    \caption{A comparison of existing model-based approaches for goal recognition (left) and our proposed framework (right). The key changes in our approach are presented in red.}
    \label{fig:cooking}
\end{figure*}

Goal recognition (GR) is a key task in artificial intelligence, where a \textit{recognizer} infers the goal of an \textit{actor} based on a sequence of observations. 
Consider a service robot that wishes to assist a person in the kitchen by fetching appropriate utensils without interrupting the task execution or demanding attention for specifying instructions \cite{kautz1986generalized,granada2020object,bishop2020chaopt}. 
A common approach to enable the robot to perceive and infer the person's goal in this situation consists of a pipeline of activity recognition from raw images and translation into actions for a symbolic GR algorithm (Figure~\ref{fig:cooking}). 
Once the raw images are processed into observations, a goal recognizer further processes a sequence of these observations into a goal or a distribution of goals. 
Most GR approaches rely on an arduous process to inform the recognizer about the feasibility and likelihood of the different actions that the actor can execute. 
This process might include crafting elaborate domain theories, multiple planner executions in real-time, intricate domain optimizations, or any combination of these tasks. 
There are several limitations to this process: 
\begin{description}
\item[Cost of Domain Description:] Crafted domain theories require deliberate design and accurate specification of domain dynamics, which is usually a process done manually by an expert. 
In highly complex environments, manual elicitation of such a model might even be impossible.
\item[Noise Susceptibility:] As specifying accurate domain dynamics is costly, many specifications are incomplete and cannot inform the recognizer about unlikely observations or partial observation sequences. 
This property makes many goal recognizers susceptible to noise. 
\item[Online Costs:] Some recognizers require costly online computations, such as multiple planner or parser executions. These computations can hinder the recognizer's real-time inference ability, especially when observations are processed incrementally and the goal of the actor needs to be re-evaluated many times throughout the plan execution.
\end{description}

We develop a framework to address these limitations by replacing manually crafted representations and online executions with model-free Reinforcement Learning (RL) techniques.
This framework performs efficient and noise-resistant GR without the need to craft a domain model and without any planner or parser executions during recognition. 

Our paper provides three key contributions. 
First, we revisit the GR problem definition to accommodate RL-based domains and develop a new framework for GR that relies on policies or utility functions derived from any model-free RL technique. 
Our framework consists of two main stages: 
Separate offline learning 
for each potential goal, and an online inference stage that compares an observed trajectory to those learned policies.  
The second contribution is a first instance of the new framework for tabular RL
using an off-the-shelf implementation of a Q-learning algorithm and three recognition measures: Accumulated utility, a modified KL-divergence, and Divergence Point. 
Finally,  
we evaluate the new framework 
on domains with partial and noisy observability. We show that even with a very short learning process, we can still accurately and robustly perform GR on challenging problems. We further demonstrate this framework's ability to perform comparably to a state-of-the-art goal recognizer on standard evaluation domains, and have superior performance in noisy environments.

\section{Background}
\label{sec:background}

We begin by defining a GR problem in a way that is consistent with existing literature \cite{Meneguzzi2021,mirsky2021introduction}. Given a domain theory $\theory$, a set of possible goals $\goals$, and a sequence of observations $\observation$, a goal recognition problem consists of a goal $\goal \in \goals$ that \textbf{$\observation$ explains}. 
The semantics of $\theory$ and \textbf{explains} can vary greatly between goal recognizers. 
For example, in Ramirez and Geffner \shortcite{ramirez2009plan}, a domain theory is a planning domain instantiated in a specific initial state $\initialstate$ and goal $\goal$ is explained by $\observation$ if there is some optimal plan for $\goal$, generated by a planner, that begins in $\initialstate$ and is compatible with $\observation$. 
They refine this interpretation to rank goals' likelihood when there is more than one goal with an optimal plan that is compatible with $\observation$ \cite{ramirez2010probabilistic}. 
In this work we propose multiple semantics for \textbf{explains}, but we first focus on defining our RL-based domain theory. 
For that, we use the definition of a Markov Decision Process (MDP), a policy, and a Q-function~\cite{sutton2018reinforcement}. 

\begin{definition}[MDP]
A \textbf{Markov Decision Process} $M$, is a 4-tuple $(\statespace, \actionspace, p, r)$ such that $\statespace$ are the possible states in the environment, $\actionspace$ is the set of actions the actor can execute, $p(s' | s, a)$ is a transition function that gives the probability of transitioning from state $s$ to state $s'$ after taking action $a$ and $r(s, a, s')$ defines a reward function.
\end{definition}

A \textbf{policy} $\pi(a \mid s)$ for a MDP is a function that defines the probability of the agent taking action $a \in \actionspace$ in state $s \in \statespace$. Some RL algorithms, such as Q-learning, compute the policy of an agent using a \textbf{Q-function} $Q(s,a)$, which is an estimation of the expected rewards of all future steps starting from $s$ after taking action $a$. %
In our new framework, a domain theory $\theory$ consists of the state and action spaces of an MDP and a set of policies or Q-functions. 
Unlike planning-based GR where the domain theory is decoupled from the problem instance (the set of possible goals $\goals$), here $\theory$ depends on the set of goals. 
We present two types of domain theories:

\begin{definition}[Utility-based Domain Theory]
A utility-based domain theory $\theory_Q(\goals)$ is a tuple $(\statespace, \actionspace, \mathcal{Q})$ such that $\mathcal{Q}$ is a set of Q-functions $\{Q_g\}_{g \in \goals}$.
\end{definition}

\begin{definition}[Policy-based Domain Theory]
A policy-based domain theory $\theory_{\pi}(\goals)$ is a tuple $(\statespace, \actionspace, \Pi)$ such that $\Pi$ is a set of policies $\{\pi_g\}_{g \in \goals}$.
\end{definition}

Essentially, both domain theories can be viewed as a set of MDPs with the same transitions but different reward functions for different goals. Our aim is to learn either a good policy or a utility function that represents the expected behavior of actors under each of these MDPs. We use this formulation to provide a new definition for a goal recognition problem in which we replace the abstract notion of $\theory$ and the combine the goal set $\goals$ into these domain theories.

\begin{definition}[Goal Recognition Problem]
Given a domain theory $\theory_Q(\goals)$ or $\theory_{\pi}(\goals)$ and a sequence of observations $\observation$, output a goal $g\in \goals$ that \textbf{$\observation$ explains}.
\end{definition}

Note that every GR approach using a policy-based domain theory $\theory_{\pi}(\goals)$, can also be given by a utility-based domain theory $\theory_Q(\goals)$. This change is performed by generating for each goal $g$ a softmax policy $\pi_g$ based on $Q_g$, as shown in Equation~\ref{eq:softmax}. Consider a case where all values in a Q-function are negative. This case would actually result in the policy $\pi_g$ being more likely to take the worst action. Thus, if the utility function has some negative values, we rescale the function with the additive inverse of the largest negative value (some number $-C$): $Q' = Q(s,a) + C$. This modification ensures that the resulting policy $\pi_g$ will prioritize high-value actions.
Thus, for brevity, from now on we refer to our framework as one that relies on Q-functions, unless we explicitly wish to discuss specific properties of a policy-based domain theory.

\begin{equation}
\label{eq:softmax}
    \pi_g(a \mid s) = \frac{Q_g(s,a)}{\sum_{a' \in \actionspace} Q_g(s,a')}
\end{equation}

Given this new problem definition, we develop our framework to solve these goal recognition problems, discuss how $\theory_Q(\goals)$ or $\theory_\pi(\goals)$ can be learned, and how to decide, once the observations $\observation$ are given, which goal $g$ they \textbf{explain} best.

\section{The Goal Recognition as Reinforcement Learning Framework}
\label{sec:framework}

\begin{figure}[t]
    \centering
    \includegraphics[width=0.45\textwidth]{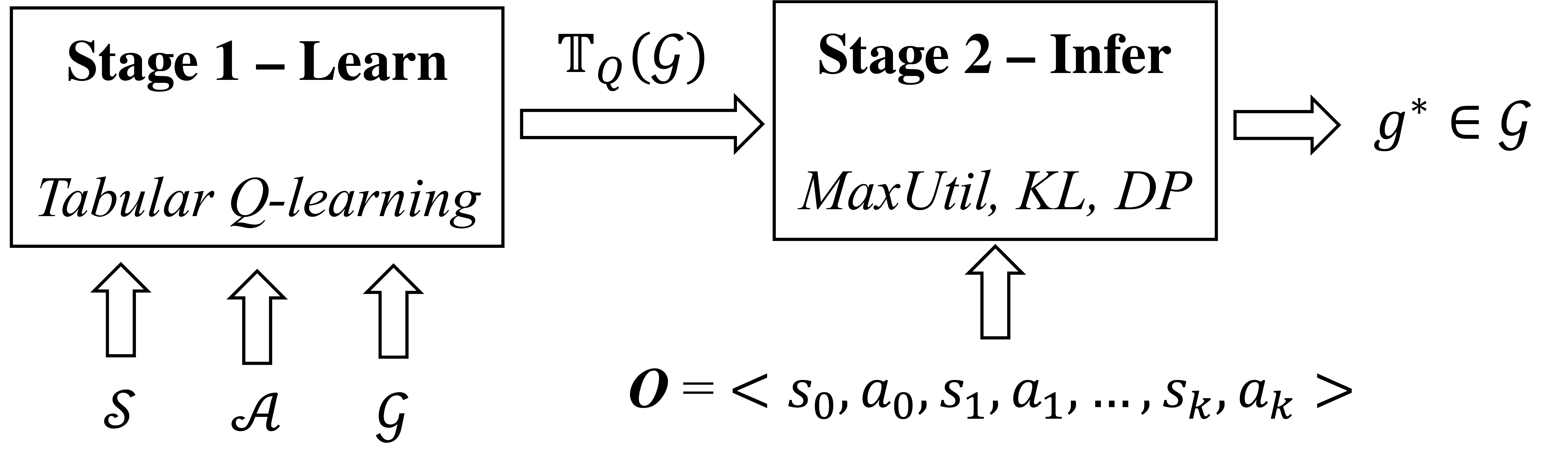}
    \caption{The proposed framework for GR as RL. Details of the Goal Recognition as Q-Learning (GRAQL) instantiation appear in \textit{italics}.}
    \label{fig:RL4GR}
\end{figure}

Our new framework consists of two main stages: (1) learning a set of Q-functions; and (2) inferring the goal of an actor given a sequence of observations. 
Figure \ref{fig:RL4GR} illustrates this process. 
First, the initial inputs are a state- and action- spaces, $\statespace$ and $\actionspace$, and a set of goals $\goals$. 
There is no restriction on the properties of $\statespace$ and $\actionspace$; they can be either discrete or continuous, and we require no transition or reward function a-priori, although they might be leveraged in specific instantiations of the framework. 
Our framework can employ any off-the-shelf RL algorithm to learn a Q-function for each goal, $\{Q_g\}_{g \in \goals}$ which together with the original $\statespace$ and $\actionspace$ constructs the domain theory $\theory_{Q}(\goals)$ for the recognition stage. 
Once the framework receives an observation sequence, a measure between an observation sequence and a Q-function computes a distance between the observations $\observation$ and each $Q_g$. 
Here we focus primarily on state-action observation sequences, where $\observation = \langle s_0, a_0, s_1, a_1, \ldots \rangle$, but we show in Section~\ref{sec:graql} examples of how to handle state-only ($\observation^s = \langle s_0, s_1,\ldots \rangle$) or action-only ($\observation^a = \langle a_0, a_1,\ldots \rangle$) observations.
The inferred goal $g^*$ is the one that minimizes the measured distance between its respective Q-function and the observations, as defined in Equation~\ref{eq:goal}. 
We move on to discuss the stages of the framework in detail.
\begin{equation}
\label{eq:goal}
    g^* = \argmin_{g \in \goals} {\text{Distance}(Q_g, \observation)}
\end{equation}

\noindent \textbf{Stage 1: Learning} The first part of the GR as RL framework is learning a set of Q-functions (or policies) for each goal, as presented in Algorithm \ref{alg:learn}. 
It defines a set of $n$ RL problems where $n$ is the number of goals in $\goals$: 
For each goal $g$, we generate a reward function in which there is some positive gain when reaching the goal $g$ (Line~\ref{alg:r_goal}), and no reward otherwise (Line~\ref{alg:r_not_goal}). 
Given this basic setting, we can leverage additional reward shaping or other optimizations to improve the learning of the Q-functions, just as in any other RL problem. 
However, in the most naive form of this problem, we require no penalty for actions that fail to advance the agent towards reaching the goal, nor any specific discount factor. 
This simple problem formulation is sufficient for our purpose, which is to generate informative Q-functions, not to perfectly maximize the reward of the RL agent. 
As we show in our empirical evaluation, even though we do not train Q-functions until convergence and provide poor solvers for reaching their respective goals, they suffice to create an accurate and robust domain theory for our goal recognizers. 

This formulation enables us to use well-established RL research to learn a set of Q-functions given the properties of our environment: discrete or continuous, deterministic or stochastic transitions, etc. 
The learning of the domain theory then becomes an RL problem with its respective challenges: selecting the most appropriate algorithm for learning, tuning its hyperparameters, etc.

\begin{algorithm}[h]
    \caption{Learn a Q-function for each goal}
    \label{alg:learn}
\begin{algorithmic}[1]
    \small
    \Require $\statespace, \actionspace:$ State and action spaces
    \Require $\goals$: a set of candidate goals
    \ForAll{$\goal \in \goals$}
        \State $\forall a \forall s \neq g, r(s, a) \gets 0$ \Comment{Create a reward function}
        \label{alg:r_not_goal}
        \State $\forall a, r(g, a) \gets C$ \Comment{Reaching $g$ yields some positive value}
        \label{alg:r_goal}
        \State $Q_g \gets$ \Call{Learn}{$\statespace, \actionspace, r$}
    \EndFor
    \State \textbf{return} $\{Q_g\}_{g \in \goals}$
\end{algorithmic}
\end{algorithm}

\noindent \textbf{Stage 2: Inference} Once we have a set of Q-functions $Q_{\goals}$ and an observation sequence $\observation$, the next stage in the framework is online GR: inferring Q-function (i.e. goal) of the actor that $\observation$ explains. 
Traditional GR algorithms require complex computations, such as planner or parser executions to reason about the similarity of $\observation$ to each goal. 
In this work, we take on a measure-based approach instead (and present potential measures in Section \ref{sec:graql}). 
Algorithm~\ref{alg:infer} implements the inference process using $\textsc{Distance}$ as a measure function, which implements Equation~\ref{eq:goal}: given $\observation$, find for each goal $g$ the distance between $\observation$ and $Q_g$ (Line \ref{alg:dist}). 
The algorithm then chooses the goal with the minimum distance value as the most likely goal of the actor (Line \ref{alg:min_dist}). 
This formulation of the GR task aligns well with Ramirez and Geffner \shortcite{ramirez2010probabilistic}, who introduce the notion of similarity between an observation sequence and optimal, goal- and observation-dependent plans. 
As the algorithm computes the similarity between the observation sequence and a Q-function that is defined over all state-action pairs rather than a single trajectory, it inherently reasons about noisy and missing observations, as shown in our empirical evaluation.

\begin{algorithm}[t]
    \caption{Infer most likely goal for the observations}
    \label{alg:infer}
\begin{algorithmic}[1]
    \small
    \Require $\theory_Q(\goals)$: $\statespace, \actionspace, \{Q_g\}_{g \in \goals}:$ State and action spaces, and Q-functions per goal
    \Require $\observation$: an observation sequence $\langle s_0, a_0, s_1, a_1, \ldots \rangle$
    \State $m_{g^*} \gets \infty$ \Comment{Init shortest distance}
    \ForAll{$\goal \in \goals$} \Comment{Compute distances from $\observation$}
        \State $m_g \gets \Call{Distance}{Q_g, \observation}$ \Comment{Use distance measure}
        \label{alg:dist}
        \If {$m_g \leq m_{g^*}$}
            \State $g^* \gets g$ and $m_{g^*} \gets m_g$
        \EndIf
    \EndFor
    \State \textbf{return} $g^*$
    \label{alg:min_dist}
\end{algorithmic}
\end{algorithm}

\section{Goal Recognition as Q-Learning}
\label{sec:graql}

In this section, we detail the components of the first instance of our framework specifically for tabular Q-learning approaches --- Goal Recognition as Q-Learning (GRAQL). 
We chose to focus the first instance of this framework on tabular domains to enable an evaluation to existing GR baselines. Planning-based GR algorithms use PDDL as their domain descriptions, which can be easily translated into tabular representation~\cite{ramirez2009plan,amado2018goal}.
Figure~\ref{fig:RL4GR} illustrates the specific components that require implementation in \textit{italics}. We start by explaining the hyper-parameters and discussing these choices in the learning stage. 
Then, we introduce three different measures for an observation sequence and a Q-function.

\subsection{Learning $\{Q_g\}_{g\in \goals}$ using Q-Learning}
For the learning stage, we use an off-the-shelf Q-learning algorithm.
As learning the Q-functions for each goal are a means to an end rather than our ultimate aim, we do not focus on techniques to optimize this stage, but rather employ a single solution, showing that we can acquire an informative domain theory with minimal effort. 
We set the reward for reaching the goal to 100, and 0 otherwise, and the discount factor to 0.9. 
As exploration is more important in this case than maximizing the reward, the sampling strategy we use is $\epsilon$-greedy with linearly decaying values ($\epsilon = 1 \ldots 0.01$). 

Shaping the initial policy can speed up the learning process: for each goal $g$, an optimal planner generates a single trajectory to the goal $p_g = \{\langle s_0, a_0 \rangle, \langle s_1, a_1 \rangle, \ldots \}$. 
We can use this trajectory to initialize $Q_g$ with positive values for state-action pairs that are part of its goal's optimal path $p_g$.
\begin{equation}
    Q_g(s,a) =
    \begin{cases}
    1,& \text{if } \langle s,a \rangle \in p_g \\
    0,              & \text{otherwise}
\end{cases}
\end{equation}

This essentially initializes it to give high utility to this single trajectory, which is similar to the original formulation of planning-based GR, where a single optimal plan constitutes the baseline for the actor's presumed path to the goal~\cite{ramirez2009plan}. 
In that sense, GRAQL bridges a gap between planning-based GR and RL: in the original formulation of Ramirez and Geffner, a planner outputs a single optimal plan for goal $g$, which might not be the plan the actor chooses to follow. 
Later work overcomes this problem by considering top-$k$ plans for each goal, by using a top-k planner that keeps $A^*$ queue running after finding the first solution~\cite{sohrabi2016plan,sohrabi2016finding}. 
Instead of this top-$k$ search, GRAQL \textbf{refines} a single optimal plan into a  policy that can capture the cost of alternative plans, even if these plans are not part of the top-$k$ ones. 
We implemented our approach with and without this shaping process in the learning stage.
To ensure that it does not overfit or create an unfair bias in the Q-functions towards the planning-based observation sequence used in our evaluation, we explicitly chose problems with multiple optimal plans per goal and ran different planners for shaping (LAMA \cite{richter2010lama}) and for testing (Fast Downward \cite{helmert2006fast}), so that $p_g$ is not the only possible optimal path. 
The resulting Q-functions with and without shaping were not significantly different, so our empirical results only show the performance of the Q-functions without the shaping process.

\subsection{Measures for Tabular $Q$-functions}
For the inference stage, we use three different measures for a distance between a Q-function $Q_g$ and an action-state observation sequence $\observation$, inspired by three common RL measures: MaxUtil, KL-divergence, and Divergence Point. 
We then extend MaxUtil's definition to handle state-only observation sequences and action-only observation sequences.

\noindent \textbf{MaxUtil} is an accumulation of the utilities collected from the observed trajectory. 

\begin{equation}
    MaxUtil(Q_g, \observation) = \displaystyle\sum\limits_{i \in |\observation|} Q_g(s_i,a_i)
\end{equation} 

\noindent \textbf{KL-Divergence} is a measure for the distance between two distributions, so we construct two policies, $\pi_g$ and $\pi_\observation$ for $Q_g$ and $\observation$ respectively. The goal-dependent policy $\pi_g$ is defined as a softmax stochastic policy as presented in Equation \ref{eq:softmax}. The observations policy $\pi_\observation$ is a pseudo-policy  where $\pi_\observation(a_i \mid s_i) = 1$ for each $\tuple{s_i, a_i} \in \observation$ and provides a uniform distribution for all actions taken in unobserved states. 
\begin{equation}
\begin{split}
    KL(Q_g, \observation) & = \kldiv(\pi_g \mid \mid \pi_\observation) = \\
    & \sum\limits_{i \in |\observation|} \pi_g(a_i \mid s_i) \log \frac{\pi_g(a_i \mid s_i)}{\pi_\observation(a_i \mid s_i)}
\end{split}
\end{equation}
\noindent \textbf{Divergence Point (DP)} is a measure adapted from Macke et al. \shortcite{AAAI21-Macke}, where given a trajectory $\observation$ and a policy $\pi$, it is defined as the minimal point in time in which the action taken by $\observation$ has zero probability to be chosen by $\pi$. 
We implement a softer version of the original measure, where the probability threshold is a parameter $\delta$ instead of exactly $0$.
The reason for this softened version of DP is that, for similar enough goals, the probability of an action to be chosen for both goals is unlikely to be exactly 0. 
Finally, as DP gets higher values when $\observation$ and $\pi$ share more resemblance, we take its additive inverse to get a distance compatible with the minimization formulation of Algorithm~\ref{alg:infer}. 
Here as well, the goal-dependent policy $\pi_g$ is defined as a softmax stochastic policy as presented in Equation \ref{eq:softmax}:
\begin{equation}
    DP(Q_g, \observation) = -min\{t \mid \pi_g(a_{t-1} \mid s_{t-1}) \leq \delta \}
\end{equation}
\noindent \textbf{MaxUtil for State-only $\observation$} applies to states-only: $\observation^s = \langle s_0, s_1, \ldots \rangle$. 
In this case, similar to offline policy learning, we optimistically take the action with the highest Q-value:
\begin{equation}
\label{eq:max_util_state}
    MaxUtil(Q_g, \observation^s) = \displaystyle \sum\limits_{i \in |\observation|} \max\limits_{a} Q_g(s_i,a)
\end{equation} 
\noindent \textbf{MaxUtil for Action-only $\observation$} applies to actions-only: $\observation^a = \langle a_0, a_1, \ldots \rangle$. 
We optimistically take the state with the highest Q-value in this case as well. To do that, we first need to find the set of all states for which the observation $a_i$ is an optimal action according to $Q_g$. 
This set can be formally defined as $Opt(a_i \mid Q_g)=\{s \in \statespace \mid Q_g(s,a_i) \geq Q_g(s,a) \forall a \in \actionspace\}$. 
From this set, we choose the state with the maximal utility (presumed to be the state in the optimal path) and associate this utility with the observation.
\begin{equation}
\label{eq:max_util_action}
    MaxUtil(Q_g, \observation^a) = \displaystyle \sum\limits_{i \in |\observation|} \max\limits_{s \in Opt(a_i \mid Q_g)} Q_g(s,a_i)
\end{equation} 

\section{Experimental Evaluation}
\label{sec:empirical}

To be able to compare GRAQL and planning-based GR, we use PDDLGym \cite{silver2020pddlgym} as our evaluation environment. 
PDDLGym is a python framework that automatically constructs OpenAI Gym environments from PDDL domains and problems. 
Thus, for each PDDL domain used by state-of-the-art GR algorithms, we generate the parallel representation in Gym for GRAQL. 
We use three domains from the PDDLGym library for their similarity with commonly used GR evaluation domains: Blocks, Hanoi, and SkGrid (The latter highly resembles common GR navigation domains such as those used by \citeauthor{masters2019cost}~\shortcite{masters2019cost}). 
For each domain, we generate $10$ GR problems with $4$ candidate goals in $\goals$. We manually choose ambiguous goals, i.e., goals that are close to one another rather than in different corners of a grid. 
Each problem has $7$ variants, including partial and noise observations. We have $5$ variants with varying degrees of observability (10\%, 30\%, 50\%, 70\%, and full observability), and $2$ variants that include noise observations with varying degrees of observability (50\% and full observability). Thus, our test set includes 210 GR problems. These GR problems can pose a real challenge to existing recognizers, especially when they are partially observable or noisy.

Next, we discuss the hyperparameters used in this evaluation and explain our three sets of performance tests, using different types of $\observation$ as input: state-action pairs where $\observation$ is fully observable; state-action pairs with missing observations, noisy observations, or both; and state-only or action-only trajectories with missing observations. 
We use standard machine learning metrics in our evaluation: accuracy, precision, recall, and F-score. 
We note that the accuracy metric reported here is different from the accuracy metric in  \citeauthor{ramirez2010probabilistic}~\shortcite{ramirez2010probabilistic}, which refers to the ratio of problems where the recognizer ranks the true goal (i.e.,  True Positives) with the highest likelihood, and the number of problems tested. 
This measure differs from standard ML accuracy, which also takes into consideration whether the algorithm successfully ranks incorrect goals \emph{lower} than the correct one (i.e., True Positives + True Negatives). 
In our experiments, we report accuracy using the latter definition.

\subsection{Hyperparameters and Algorithmic Choices}
For the learning stage of our experiments, we use an off-the-shelf model-free Q-learning algorithm~\cite{Sutton1988}. 
For each goal, we run the learner for a fixed number of episodes, whether it reaches convergence or not. 
We evaluate greedy policy executions after training the Q-functions for $500$, $10k$ and $30k$ episodes. The difference in the resulting plan lengths was around $5\%$, and went both ways---sometimes the policy learned after $500$ episodes was not informed enough and drifted away from the goal during execution, and sometimes policies learned after $30k$ episodes overfitted a specific solution. 
The key differentiator is often whether the learner reaches the goal once during training, as after that the learned Q-function can generate an optimal plan for that goal. 
Since the performance of these training regimes does not vary much, we report our empirical results with a consistent value of $500$ training episodes.

For the inference stage, we compare our framework with each of our three proposed measures, denoted MaxUtil, KL, and DP, as well as the probabilistic planning-based recognizer by Ramirez and Geffner \shortcite{ramirez2010probabilistic}, denoted R\&G. 
For the DP measure, we use a threshold probability for divergence of $\delta=0.1$. As we wish to be able to compare GRAQL against a planning-based algorithm, we focus our evaluation on discrete state and action spaces, but we conjecture that the framework can directly handle continuous spaces by using the same measures, only with a different learning algorithm. 

\subsection{Testing under Full Observability}

\begin{figure}[t]
    \centering
    \includegraphics[width=0.4\textwidth]{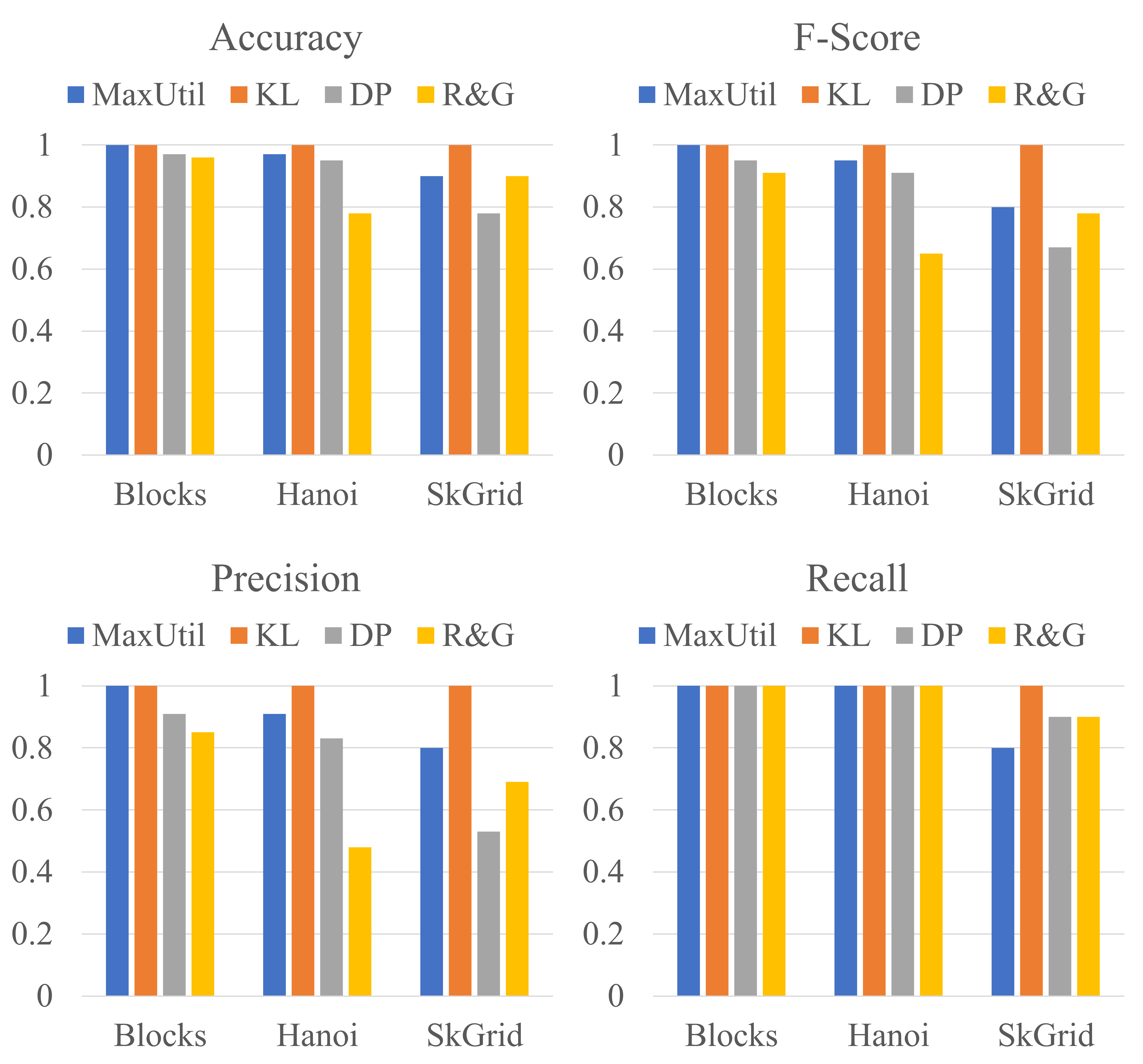}
    \caption{Comparison of R\&G, MaxUtil, KL, DP by their accuracy, precision, recall, and F-score for full observability.}
    \label{fig:full_observability}
\end{figure}

First, we show the performance of MaxUtil, KL, DP, and R\&G on fully observable, optimal state-action trajectories. 
The average length of these trajectories is $8$ $(\min=5, \max=15)$ steps for Blocks, $10.4$ $(\min=5, \max=15)$ for Hanoi, and $12.4$ $(\min=6, \max=24)$ for SkGrid. 
Figure~\ref{fig:full_observability} summarizes the performance of each approach over ten instances per setup. 
With full observability, almost all results are above $80\%$, and KL achieves perfect performance, consistently predicting the actor's goal correctly. 
Notably, DP performs worse than the other GRAQL measures, especially in the SkGrid environment. As DP requires an action to be highly unlikely for some goal to rank it as incorrect, the high variability of SkGrid presents a real challenge to it. In this domain, the same goal might have a number of equally optimal trajectories requiring diagonal movement due to the nature of square-grid navigation environments. 
All GRAQL approaches perform similarly or better than R\&G, except for DP on SkGrid, where the largest difference was in precision (DP: $0.53$, R\&G: $0.69$). 
On the other hand, the performance of R\&G in the Hanoi environment is inferior to all GRAQL methods in terms of accuracy, (DP: $0.95$, R\&G: $0.78$), precision (DP:$0.83$, R\&G: $0.48$), and F-score (DP:$0.91$, R\&G:$0.65$). Hanoi has many actions that appear in plans for different goals, causing high ambiguity in recognition time, which makes it especially challenging to R\&G to distinguish between those goals.
These results show that GRAQL is able to achieve comparable results to the state-of-the-art with fully observable trajectories.

\subsection{Testing under Partial Observability and Noise}

\begin{table*}[htb]
    \caption{Impact of partial observability: comparing MaxUtil, KL, DP, R\&G with varying observability levels of $\observation$.}
    \label{tab:partial_observability}
    \tiny
    \begin{tabular}{cc||l|l|l|l||l|l|l|l||l|l|l|l||l|l|l|l|}
    \cline{3-18}
                                               & \multicolumn{1}{c|}{}       & \multicolumn{4}{c||}{Accuracy}                                                                                    & \multicolumn{4}{c||}{Precision}                                                                                   & \multicolumn{4}{c||}{Recall}                                                                                      & \multicolumn{4}{c|}{F-Score}                                                                                     \\ \hline
    \multicolumn{1}{|c|}{OBS}                  & \multicolumn{1}{c||}{Domain} & \multicolumn{1}{c|}{MaxUtil} & \multicolumn{1}{c|}{KL Div} & \multicolumn{1}{c|}{DP} & \multicolumn{1}{c||}{R\&G} & \multicolumn{1}{c|}{MaxUtil} & \multicolumn{1}{c|}{KL Div} & \multicolumn{1}{c|}{DP} & \multicolumn{1}{c||}{R\&G} & \multicolumn{1}{c|}{MaxUtil} & \multicolumn{1}{c|}{KL Div} & \multicolumn{1}{c|}{DP} & \multicolumn{1}{c||}{R\&G} & \multicolumn{1}{c|}{MaxUtil} & \multicolumn{1}{c|}{KL Div} & \multicolumn{1}{c|}{DP} & \multicolumn{1}{c|}{R\&G} \\ \hline
    \multicolumn{1}{|c|}{\multirow{3}{*}{10\%}} & Blocks                      & \textbf{0.93}                & 0.90                        & \textbf{0.93}           & 0.44                      & \textbf{0.82}                & 0.80                        & 0.77                    & 0.25                      & 0.90                         & 0.80                        & \textbf{1.00}           & 0.90                      & 0.86                         & 0.80                        & \textbf{0.87}           & 0.39                      \\ \cline{2-18} 
    \multicolumn{1}{|c|}{}                     & Hanoi                       & 0.93                         & \textbf{0.95}               & 0.90                    & 0.50             & 0.77                         & \textbf{0.90}               & 0.71                    & 0.29                      & \textbf{1.00}                & 0.90                        & \textbf{1.00}           & \textbf{1.00}             & 0.87                         & \textbf{0.90}               & 0.83                    & 0.44                      \\ \cline{2-18} 
    \multicolumn{1}{|c|}{}                     & SkGrid                      & 0.80                         & \textbf{0.90}               & 0.55                    & 0.72             & 0.60                         & \textbf{0.80}               & 0.36                    & 0.42                      & 0.60                         & 0.80                        & \textbf{1.00}           & \textbf{1.00}             & 0.60                         & \textbf{0.80}               & 0.53                    & 0.59                      \\ \hline
    \multicolumn{1}{|c|}{\multirow{3}{*}{30\%}} & Blocks                      & \textbf{1.00}                & 0.95                        & 0.97                    & 0.74                      & \textbf{1.00}                & 0.90                        & 0.91                    & 0.42                      & \textbf{1.00}                & 0.90                        & \textbf{1.00}           & 0.80                      & \textbf{1.00}                & 0.90                        & 0.95                    & 0.55                      \\ \cline{2-18} 
    \multicolumn{1}{|c|}{}                     & Hanoi                       & \textbf{0.95}                & \textbf{0.95}               & 0.93                    & 0.68             & 0.83                         & \textbf{0.90}               & 0.77                    & 0.38                      & \textbf{1.00}                & 0.90                        & \textbf{1.00}           & \textbf{1.00}             & \textbf{0.91}                & 0.90                        & 0.87                    & 0.55                      \\ \cline{2-18} 
    \multicolumn{1}{|c|}{}                     & SkGrid                      & 0.90                         & \textbf{0.95}               & 0.70                    & 0.88             & 0.80                         & \textbf{0.90}               & 0.45                    & 0.63                      & 0.80                         & 0.90                        & \textbf{1.00}           & \textbf{1.00}             & 0.80                         & \textbf{0.90}               & 0.62                    & 0.77                      \\ \hline
    \multicolumn{1}{|c|}{\multirow{3}{*}{50\%}} & Blocks                      & \textbf{1.00}                & \textbf{1.00}               & 0.97                    & 0.80                      & \textbf{1.00}                & \textbf{1.00}               & 0.91                    & 0.50                      & \textbf{1.00}                & \textbf{1.00}               & \textbf{1.00}           & 0.90                      & \textbf{1.00}                & \textbf{1.00}               & 0.95                    & 0.64                      \\ \cline{2-18} 
    \multicolumn{1}{|c|}{}                     & Hanoi                       & \textbf{0.95}                & \textbf{0.95}               & 0.93                    & 0.72             & 0.83                         & \textbf{0.90}               & 0.77                    & 0.42                      & \textbf{1.00}                & 0.90                        & \textbf{1.00}           & \textbf{1.00}             & \textbf{0.91}                & 0.90                        & 0.87                    & 0.59                      \\ \cline{2-18} 
    \multicolumn{1}{|c|}{}                     & SkGrid                      & 0.80                         & \textbf{0.90}               & 0.72                    & 0.88             & 0.60                         & \textbf{0.80}               & 0.48                    & 0.63                      & 0.60                         & 0.80                        & \textbf{1.00}           & \textbf{1.00}             & 0.60                         & \textbf{0.80}               & 0.65                    & 0.77                      \\ \hline
    \multicolumn{1}{|c|}{\multirow{3}{*}{70\%}} & Blocks                      & \textbf{1.00}                & \textbf{1.00}               & 0.97                    & 0.94             & \textbf{1.00}                & \textbf{1.00}               & 0.91                    & 0.77                      & \textbf{1.00}                         & \textbf{1.00}               & \textbf{1.00}           & \textbf{1.00}             & \textbf{1.00}                & \textbf{1.00}               & 0.95                    & 0.87                      \\ \cline{2-18} 
    \multicolumn{1}{|c|}{}                     & Hanoi                       & \textbf{0.95}                & 0.90                        & 0.93                    & 0.72             & \textbf{0.83}                & 0.80                        & 0.77                    & 0.42                      & \textbf{1.00}                         & 0.80                        & \textbf{1.00}           & \textbf{1.00}             & \textbf{0.91}                & 0.80                        & 0.87                    & 0.59                      \\ \cline{2-18} 
    \multicolumn{1}{|c|}{}                     & SkGrid                      & 0.85                         & \textbf{0.95}               & 0.72                    & 0.92             & 0.70                         & \textbf{0.90}               & 0.47                    & 0.71                      & 0.70                         & 0.90                        & 0.90                    & \textbf{1.00}             & 0.70                         & \textbf{0.90}               & 0.62                    & 0.83                      \\ \hline
    \multicolumn{1}{|c|}{\multirow{3}{*}{100\%}}   & Blocks                      & \textbf{1.00}                & \textbf{1.00}               & 0.97                    & 0.96             & \textbf{1.00}                & \textbf{1.00}               & 0.91                    & 0.85                      & \textbf{1.00}                         & \textbf{1.00}               & \textbf{1.00}           & \textbf{1.00}             & \textbf{1.00}                & \textbf{1.00}               & 0.95                    & 0.92                      \\ \cline{2-18} 
    \multicolumn{1}{|c|}{}                     & Hanoi                       & 0.97                         & \textbf{1.00}               & 0.95                    & 0.78             & 0.91                         & \textbf{1.00}               & 0.83                    & 0.48                      & \textbf{1.00}                         & \textbf{1.00}               & \textbf{1.00}           & \textbf{1.00}             & 0.95                         & \textbf{1.00}               & 0.91                    & 0.65                      \\ \cline{2-18} 
    \multicolumn{1}{|c|}{}                     & SkGrid                      & 0.90                         & \textbf{1.00}               & 0.78                    & 0.90                      & 0.80                         & \textbf{1.00}               & 0.53                    & 0.69                      & 0.80                         & \textbf{1.00}               & 0.90                    & 0.90                      & 0.80                         & \textbf{1.00}               & 0.67                    & 0.78                      \\ \hline \hline
    \multicolumn{1}{|c|}{\multirow{3}{*}{Avg}} & Blocks                      & \textbf{0.98}                & 0.97                        & 0.96                    & 0.78                      & \textbf{0.96}                & 0.94                        & 0.88                    & 0.56                      & 0.98                         & 0.94                        & \textbf{1.00}           & 0.92                      & \textbf{0.97}                & 0.94                        & 0.93                    & 0.67                      \\ \cline{2-18} 
    \multicolumn{1}{|c|}{}                     & Hanoi                       & \textbf{0.95}                & \textbf{0.95}               & 0.93                    & 0.68             & 0.83                         & \textbf{0.90}               & 0.77                    & 0.40                      & \textbf{1.00}                         & 0.90                        & \textbf{1.00}           & \textbf{1.00}             & \textbf{0.91}                & 0.90                        & 0.87                    & 0.56                      \\ \cline{2-18} 
    \multicolumn{1}{|c|}{}                     & SkGrid                      & 0.85                         & \textbf{0.94}               & 0.69                    & 0.86                      & 0.70                         & \textbf{0.88}               & 0.45                    & 0.62                      & 0.70                         & 0.88                        & 0.96                    & \textbf{0.98}             & 0.70                         & \textbf{0.88}               & 0.61                    & 0.75                      \\ \hline
    \end{tabular}
    \end{table*}

We evaluate our approaches under partial observability with varying degrees of observability (10\%, 30\%, 50\%, 70\%, and full observability). 
We use the same trajectories in all experiments, removing steps randomly to achieve a specific observability ratio. 
Table~\ref{tab:partial_observability} shows the average performance of each approach over ten instances per setup. 
It is clear that as observability decreases, so does the performance of all approaches. 
However, partial observability seems to highly affect the performance of R\&G, with values that decrease to about a half in the $10\%$ observability level (e.g., accuracy of $0.96$ drops to $0.44$ in Blocks). 
In general, KL and MaxUtil perform better than DP and R\&G, except for the recall metric, where these latter ones are the superior approaches. 
The reason for this loss of recall is that DP and R\&G are more likely to have ties between potential goals, so when these approaches cannot distinguish between them using $\observation$, they return multiple goals, trading off recall with precision.

 \begin{table*}[tb]
        \caption{Impact of noise: comparing MaxUtil, KL, DP, R\&G  with varying observability and with 4 noisy observations in $\observation$.}
        \label{tab:noisy_observations}
        \tiny
        \begin{tabular}{cl|r|r|r|r|r|r|r|r|r|r|r|r|r|r|r|r|}
        \cline{3-18}
                                                   & \multicolumn{1}{c|}{} & \multicolumn{4}{c|}{Accuracy}                                                                                    & \multicolumn{4}{c|}{Precision}                                                                                   & \multicolumn{4}{c|}{Recall}                                                                                      & \multicolumn{4}{c|}{F-Score}                                                                                     \\ \hline
        \multicolumn{1}{|c|}{OBS}                  & Domain                & \multicolumn{1}{c|}{MaxUtil} & \multicolumn{1}{c|}{KL Div} & \multicolumn{1}{c|}{DP} & \multicolumn{1}{c|}{R\&G} & \multicolumn{1}{c|}{MaxUtil} & \multicolumn{1}{c|}{KL Div} & \multicolumn{1}{c|}{DP} & \multicolumn{1}{c|}{R\&G} & \multicolumn{1}{c|}{MaxUtil} & \multicolumn{1}{c|}{KL Div} & \multicolumn{1}{c|}{DP} & \multicolumn{1}{c|}{R\&G} & \multicolumn{1}{c|}{MaxUtil} & \multicolumn{1}{c|}{KL Div} & \multicolumn{1}{c|}{DP} & \multicolumn{1}{c|}{R\&G} \\ \hline
        \multicolumn{1}{|c|}{\multirow{3}{*}{50\%}} & Blocks                & \textbf{0.95}                & 0.62                        & 0.93                    & 0.84                      & \textbf{0.95}                & 0.33                        & 0.77                    & 0.56                      & 0.90                         & 0.50                        & \textbf{1.00}           & \textbf{1.00}             & \textbf{0.90}                & 0.40                        & 0.87                    & 0.71                      \\ \cline{2-18} 
        \multicolumn{1}{|c|}{}                     & Hanoi                 & \textbf{0.97}                & 0.90                        & 0.93                    & 0.68                      & \textbf{0.91}                & 0.80                        & 0.77                    & 0.38                      & \textbf{1.00}                & 0.80                        & \textbf{1.00}           & \textbf{1.00}             & \textbf{0.95}                & 0.80                        & 0.87                    & 0.56                      \\ \cline{2-18} 
        \multicolumn{1}{|c|}{}                     & SkGrid                & 0.75                         & 0.75                        & 0.57                    & \textbf{0.88}             & 0.50                         & 0.50                        & 0.35                    & \textbf{0.64}             & 0.50                         & 0.50                        & 0.80                    & \textbf{0.90}             & 0.50                         & 0.50                        & 0.48                    & \textbf{0.75}             \\ \hline
        \multicolumn{1}{|c|}{\multirow{3}{*}{100\%}}   & Blocks                & \textbf{1.00}                & \textbf{1.00}               & 0.95                    & 0.96                      & \textbf{1.00}                & \textbf{1.00}               & 0.83                    & 0.83                      & \textbf{1.00}                & \textbf{1.00}               & \textbf{1.00}           & \textbf{1.00}             & \textbf{1.00}                & \textbf{1.00}               & 0.91                    & 0.91                      \\ \cline{2-18} 
        \multicolumn{1}{|c|}{}                     & Hanoi                 & \textbf{1.00}                & \textbf{0.95}               & 0.90                    & 0.78                      & \textbf{1.00}                & 0.90                        & 0.71                    & 0.48                      & \textbf{1.00}                & 0.90                        & \textbf{1.00}           & \textbf{1.00}             & \textbf{1.00}                & 0.90                        & 0.83                    & 0.65                      \\ \cline{2-18} 
        \multicolumn{1}{|c|}{}                     & SkGrid                & 0.85                         & \textbf{0.95}               & 0.65                    & 0.90                      & 0.70                         & \textbf{0.90}               & 0.40                    & 0.69                      & 0.70                         & \textbf{0.90}               & 0.80                    & \textbf{0.90}             & 0.70                         & \textbf{0.90}               & 0.53                    & 0.78                      \\ \hline
        \multicolumn{1}{|c|}{\multirow{3}{*}{Avg}} & Blocks                & \textbf{0.97}                & 0.81                        & 0.94                    & 0.90                      & \textbf{0.97}                & 0.60                        & 0.80                    & 0.70                      & 0.95                         & 0.75                        & \textbf{1.00}           & \textbf{1.00}             & \textbf{0.95}                & 0.67                        & 0.89                    & 0.81                      \\ \cline{2-18} 
        \multicolumn{1}{|c|}{}                     & Hanoi                 & \textbf{0.99}                & 0.93                        & 0.91                    & 0.73                      & \textbf{0.95}                & 0.85                        & 0.74                    & 0.43                      & \textbf{1.00}                & 0.85                        & \textbf{1.00}           & \textbf{1.00}             & \textbf{0.98}                & 0.85                        & 0.85                    & 0.61                      \\ \cline{2-18} 
        \multicolumn{1}{|c|}{}                     & SkGrid                & 0.80                         & 0.85                        & 0.61                    & \textbf{0.89}             & 0.60                         & \textbf{0.70}               & 0.37                    & 0.67                      & 0.60                         & 0.70                        & 0.80                    & \textbf{0.90}             & 0.60                         & 0.70                        & 0.51                    & \textbf{0.77}             \\ \hline
        \end{tabular}
        \end{table*}

Finally, we evaluate our approaches with the addition of noise in the observations.
We add noise to the observations by first generating an optimal plan using the Fast Downward planner. 
We then randomly choose a step-index and inject two consecutive non-optimal actions from that step, thus forcibly deviating the observed plan from the optimal. 
Finally, to get back to the goal with no additional noise, we rerun the planner to get an optimal plan to the goal from the state reached after executing these two noisy actions. 
Using this process to add noise, we have four additional noisy actions per trajectory for all of our environments, two that make the agent drift away from its goal, and two additional actions to backtrack. We chose this form of noise as these actions are still valid, even if not optimal. An alternative noise could have been an injection of invalid actions, or any random state-action pair that is not part of the generated trajectory. However, R\&G and most other planning-based approaches cannot trivially reason about this type of noise, as they will simply label that noisy plan as an impossible plan for the goal. 
We tested the noisy trajectories with full observability and with partial observability set to $0.5$. 
Table \ref{tab:noisy_observations} shows these results. 
Unlike the noiseless case, in these results MaxUtil outperforms KL in most setups in terms of accuracy, precision, and F-score.
When comparing the overall performance of each approach with and without noise, KL and R\&G are more noise-sensitive than MaxUtil and DP. 

\subsection{State-Only and Action-Only Observations}

\begin{figure}[!b]
    \centering
    \includegraphics[width=0.4\textwidth]{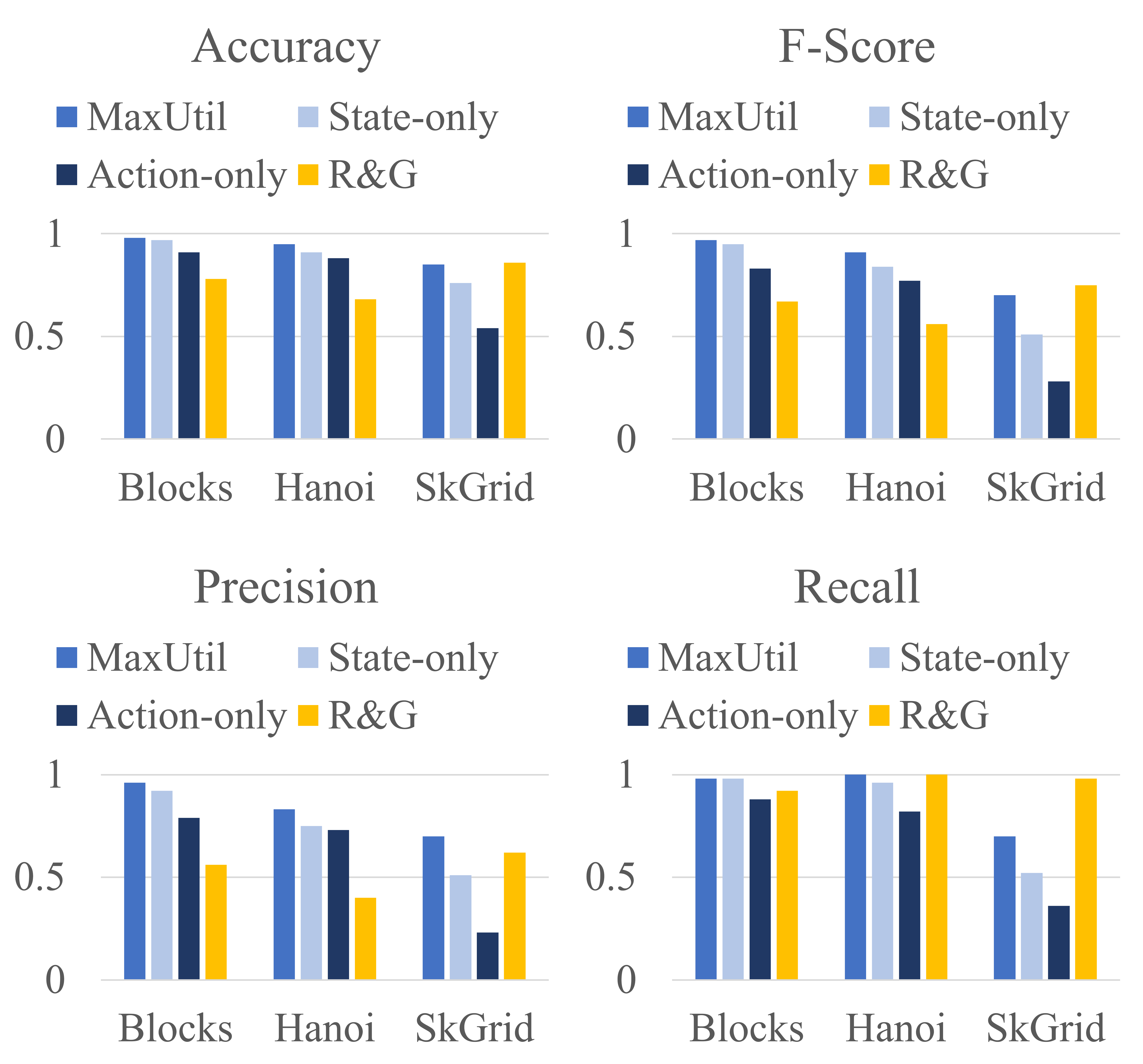}
    \caption{Performance comparison of R\&G and MaxUtil with $\observation^s$ (state-only), $\observation^a$ (action-only), and $\observation$ (state-action).}
    \label{fig:simpler_obs}
\end{figure}

R\&G, for example, uses only actions in its inference process, but no states (or vice versa), even if they are available. 
Our next set of experiments shows the performance of GRAQL when given such observations, and compares it with R\&G. 
We use the same observation sequences as in our partial observability experiments, but we provide our MaxUtil-based approaches with either the full sequences, the states $\observation^s = \langle s_0, s_1, \ldots \rangle$ (MaxUtil for state-only, in Equation \ref{eq:max_util_state}) or the actions $\observation^a = \langle a_0, a_1, \ldots \rangle$ (MaxUtil for action-only, in Equation \ref{eq:max_util_action}). 
Figure \ref{fig:simpler_obs} summarizes these results, where each bar represents the average performance for all observation levels (from $10\%$ to full). 
All versions of MaxUtil perform well in Blocks and Hanoi and outperform R\&G in all metrics but recall. 
The main issue is the Action-only version in the SkGrid domain (e.g. accuracy of $0.54$ and precision of $0.23$), which underperforms significantly against other versions. 
Equation \ref{eq:max_util_action} estimates states using the most optimistic of all of the states for which the observed action is an optimal action. 
Given SkGrid's structure, every action is an optimal action for about half of the states, thus taking this optimistic approach is unlikely to be accurate.

\section{Related Work}
\label{sec:related_work}

A large body of work involves learning for \textit{planning} domains~\cite{zimmerman2003learning,arora2018review}. 
While some approaches learn action models from data, they do not link these action models to policies for reaching specific goals~\cite{amir2008learning,amado2019latrec,asai2020learning,juba2021safe}. 
For example, Zeng et al. \shortcite{zeng2018inverse} use inverse reinforcement learning (IRL) to learn the rewards of the actor and then use an MDP-based GR. 
However, for GR, the motivation (rewards) that lead the actor to choose one action over another is redundant. 
By directly using RL, we skip this stage and learn utility functions or policies based on past actor experiences towards achieving specific goals. 
Amado et al. \shortcite{amado2018goal} learn domain theories for GR using autoencoders. 
However, they require observation of all possible transitions of a domain in order to infer its encoding, whereas we need only a small part of the transitions to learn a utility function informative enough to carry out GR effectively. 

Unlike approaches that learn models for planning, we do not reason about the plan of the acting agent, but rather about the plan of another agent. 
In this case, we cannot control the actor's choices, and we might not know or care how the actor represents the environment and the task. 
Nevertheless, we need to be able to find a good-enough explanation for its actions to be able to assist it (as in the kitchen example from Figure~\ref{fig:cooking}). 
This setup is not the one used in existing work on learning other agent's behavior, e.g., the LOPE system \cite{garcia2000integrated}, \cite{safaei2007incremental}, and IRALe \cite{rodrigues2011active}, as these systems choose the execution sequences it learns from. 
We can, however, use observed actions of other agents to improve our learning process. 
Gil \shortcite{gil1994learning} does so by investigating cases where executing new experiments can refine operators. 

Other metric-based GR use distance metrics between an optimal plan and the observation sequence, which can somewhat alleviate the need in online planner executions \cite{masters2017cost,mirsky2019new}. This work differs from this problem statement, as it relies on the distance between a Q-function and an observation sequence rather than an optimal plan and an observation sequence.

\section{Discussion and Conclusion}
\label{sec:discussion}
In this paper, we introduce a new framework for Goal Recognition (GR) as model-free reinforcement learning, which obviates the need for an explicit model of the environment and candidate goals in the GR process. 
Our framework uses learned Q-values implicitly representing the agents under observation in lieu of explicit goals from traditional GR. 
This approach allows us to solve GR problems by minimizing the distance between an 
observation sequence and Q-values representing goal hypotheses or policies extracted from them. 
We instantiate this framework in \textit{Goal-Recognition As Q-Learning} (GRAQL), which uses off-the-shelf tabular Q-learning to learn both the environmental dynamics and preferences of the agents under observation. 
This instantiation includes several possible distance measures we can derive from the Q-tables, including the difference between the stochastic policies induced by the observations and a softmax derived from the Q-tables (KL-divergence), a soft-divergence point between the observed trajectory and possible trajectories that the softmax policy could generate, and various Q-value maximization measures (MaxUtil). 
Our distance measures are competitive with the reference approach from the literature \cite{ramirez2009plan} in all experimental environments, and some distance measures outperform the reference approach in most domains, especially when the observation sequence is noisy or partial. 
However, not all approaches are uniformly superior. For instance, KL-divergence performs well in most tested cases, while MaxUtil performs better in the noisy setting. 

Besides recognition performance, GR needs to be computationally efficient so that an observer can quickly make decisions in response to the recognized goal in real-time. 
In this respect, our approaches differ substantially from recent planning-based GR, as it shifts almost all computation load to a pre-processing stage, instead of costly online planner runs. It is the case that computing the policies for each candidate goal is even more costly than running a planner for each goal. However, this computation can be done once prior to the recognition stage, and then the computation of processing observations is trivial and proportional to the number of observations. 
In addition, learning the policies saves the time of a domain expert who needs to carefully design the planning dynamics---a cost which is even harder to quantify. 

In closing, our work paves the way for a new class of GR approaches based on model-free reinforcement learning. 
Future work will focus on new, more robust distance measures and mechanisms to handle noise explicitly, as well as experimenting with models learned using function approximation (e.g., neural networks). 
While our work is theoretically compatible with non-tabular representations of the value functions (as long as the action-space is discrete), we chose to focus our experiments on domains that are translatable to PDDL so our approach can be compared to planning-based GR.
We plan to extend this work to image-based domains rather than PDDL-based ones, which will require substantial effort to build suitable GR benchmarks.

\bibliographystyle{aaai}
\bibliography{bibliography}

\end{document}